\documentclass[english]{IOS-Book-Article}
\usepackage[T1]{fontenc}
\usepackage[latin9]{inputenc}
\usepackage{units}
\usepackage{graphicx}
\usepackage[numbers]{natbib}
\usepackage{nicefrac}

\makeatletter

\providecommand{\tabularnewline}{\\}

\usepackage{times} 
\normalfont
\usepackage[T1]{fontenc}
\usepackage {url}

\@ifundefined{showcaptionsetup}{}{%
 \PassOptionsToPackage{caption=false}{subfig}}
\usepackage{subfig}
\makeatother

\usepackage{babel}
\begin{document}
\begin{frontmatter}              
\title{Support vector machine classification of dimensionally reduced structural MRI images for dementia}
\runningtitle{Classification of dimensionally Reduced MRI for dementia}

\author[A]{\fnms{Victor A.} \snm{Miller}%
\thanks{Corresponding Author e-mail: vamiller@stanford.edu}},
\author[A]{\fnms{Stephen} \snm{Erlien}}
and \author[B]{\fnms{Jeff} \snm{Piersol}}

\address[A]{Department of Mechanical Engineering, Stanford University, Stanford, CA, USA} 
\address[B]{Department of Electrical Engineering, Stanford University, Stanford, CA, USA}

\begin{abstract}. 
We classify very-mild to moderate dementia in patients (CDR ranging from $0$ to $2$) using a support vector machine classifier acting on dimensionally reduced feature set derived from MRI brain scans of the $416$ subjects available in the OASIS-Brains dataset. We use image segmentation and principal component analysis to reduce the dimensionality of the data.  Our resulting feature set contains 11 features for each subject.  Performance of the classifiers is evaluated using $10$-fold cross-validation.  Using linear and (gaussian) kernels, we obtain a training classification accuracy of $86.4\%$ ($90.1\%$), test accuracy of $85.0\%$ ($85.7\%$), test precision of $68.7\%$ ($68.5\%$), test recall of $68.0\%$ ($74.0\%$), and test Matthews correlation coefficient of $0.594$ ($0.616$).

\end{abstract}
\begin{keyword} 
support vector machine\sep CDR\sep dementia classification\sep principal component analysis\sep dimesional reduction \sep eigenbrain 
\end{keyword} 
\end{frontmatter}
\thispagestyle{empty} 
\pagestyle{empty}

\section{Introduction}

This is a technical note describing the implementation of linear and non-linear kernel
support vector machine (SVM) classifiers to classify MRI scans according
to whether or not the patient has a non-zero clinical dementia rating
(CDR). We train and test these classifiers on the Open Access Imaging
Series (OASIS) dataset \citep{Marcus2007}, which contains $416$
subjects, of which $100$ have been diagnosed with dementia (clinical
dementia rating, CDR > 0). 

Previous works have discussed applications of machine learning techniques
to classify MRI scans for dementia or Alzheimer's and depression:
Kloppel et al. \citep{Kloppel2008} utilized a support vector machine
and structural MRI images to separate healthy subjects from those
diagnosed with Alzheimer's disease (AD) via either neuropathology
or Mini Mental State Examination (MMSE) score, and the same work reported
to be successful in aiding the differential diagnosis of AD and frontotemporal
lobar degeneration. Recent work by Mwangi et al. \citep{Mwangi2012}
has also been successful in identifying depression in subjects from
structural MRI images. These previous works utilized relatively small
samples of subjects, testing and training support vector machines
on relatively small samples containing fewer than $100$ subjects.
Additionally, these works used the raw image data to train and test
the classifiers, that is, raw voxel intensity values were fed into
the algorithms. An MRI scan with $256^{3}$ voxels has over $15$
million voxels. When replicating the procedure outline in previous
works by training a linear-kernel SVM classifier using a subset ($10\%$)
of the raw voxel intensities, we observe severe over fitting of the
classifier (i.e., training accuracy of $100\%$), which may introduce
significant bias if applied out of sample. 

Therefore, in this work, we aim to improve the generality of the classifier
by reducing dimensionality of the feature set. Rather than testing
and training on the entire imaged volume, we reduce the dimensionality
of the feature set to only $11$ features per example by using quantities
computed from the images and principal component analysis (PCA), then
we assess the performance of the classifiers for various sets of features
via 10-fold cross-validation. Dimensional reduction of the data also
increases the computational tractability of the problem, enabling
one to execute the classifier on a standard laptop computer. 

This brief is structured as follows: first, we present a brief overview
of a SVM classifier and its implementation; second, we present the
data used in the study; third, we summarize and motivate the feature
set on which we train and test the classifiers; lastly, we present
and discuss our results for classifying mild- to moderate-dementia
using a dimensionally reduced dataset.

\section{Support Vector Machine}

We implement a SVM classifiers to classify subjects into a CDR score
moderate dementia using MRI scans. An SVM is able to determine an
optimal boundary between to datasets described by features via a linear
and non-linear combinations of features; for a thorough review of
support vector machines in the context of classification for diagnosis,
please see \citep{Kloppel2008} and \citep{Mwangi2012}. We use $LIBSVM$
\citep{Chang2011} to implement linear- and radial basis function
(i.e., gaussian) SVM classifiers.

\section{MRI dataset and features}

OASIS \citep{Marcus2007} provides T1-weighted MRI scans for 416 subjects,
aged 18 to 96, including both women and men, and all subjects are
right handed. Subjects' cognitive health has been diagnosed via both
mini-mental state exam (MMSE, \citep{Rubin1998}) and clinical dementia
rating (CDR, \citep{Morris1993}), and 100 subjects have a CDR greater
than $0$, with $70$ subjects having a CDR of $0.5$, $28$ with
a rating of $1$, and $2$ with a rating $2$; none have scores greater
than 2. The dataset also includes the subjects' age, gender, estimated
total intracranial volume (eTIV, \citep{Buckner2004}), and normalized
whole brain volume (nWBV, \citep{Fotenos2005}).

For more details about the data acquisition, content, and processing,
we refer you to \citep{Marcus2007}; a brief synopsis is provided
here. For every subject, OASIS provides post-processed images: the
non-brain tissue is masked using the fMRIDC Brain Extraction Tool,
and the images are gain-field corrected and registered to the 1988
atlas of Talairach and Tournoux \citep{Buckner2004}. Lastly, using
the atlas registered image, a segmented (grey/white/CSF) brain is
computed \citep{Zhang2001}. In this work, we use the masked/gain-field
corrected/registered images (referred to as 'masked' images), and
the segmented images (referred to as 'segmented' images). In the registration,
all images are resampled with $1\, mm$ isotropic voxels resulting
in a final imaged volume of $176\times208\times176$ voxels \citep{Marcus2007}.

We compute a reduced set of features using data provided in the OASIS-Brains
dataset, as well as from the masked and segmented images. We computed
a measure of symmetry $\Psi$ for a scan $I$ using the masked images;
we compute the zero-lag ($\tau=0$) correlation ($\star$) of each
slice with a flipped (left-right or up-down, noted as $I^{'}$) version
of itself, and normalizing by the integrated level of signal in that
slice

\begin{equation}
\Psi=\frac{(I_{slice}\star I_{slice}^{'})(\tau=0)}{\sum I_{slice}}.\label{eq:sym_eq}
\end{equation}
The average value of the symmetry over all axial slices is used as
the feature for a given scan. We also compute the sum of white matter,
grey matter, and CSF as the sum of the indicator function ($1_{E}$)
of $E=3$, $2$, and $1$, respectively, with the segmented images. 

\begin{figure}
\centering{}\subfloat[Coronal raw image, segmented image, and eigenbrain \#4]{\includegraphics[width=4cm]{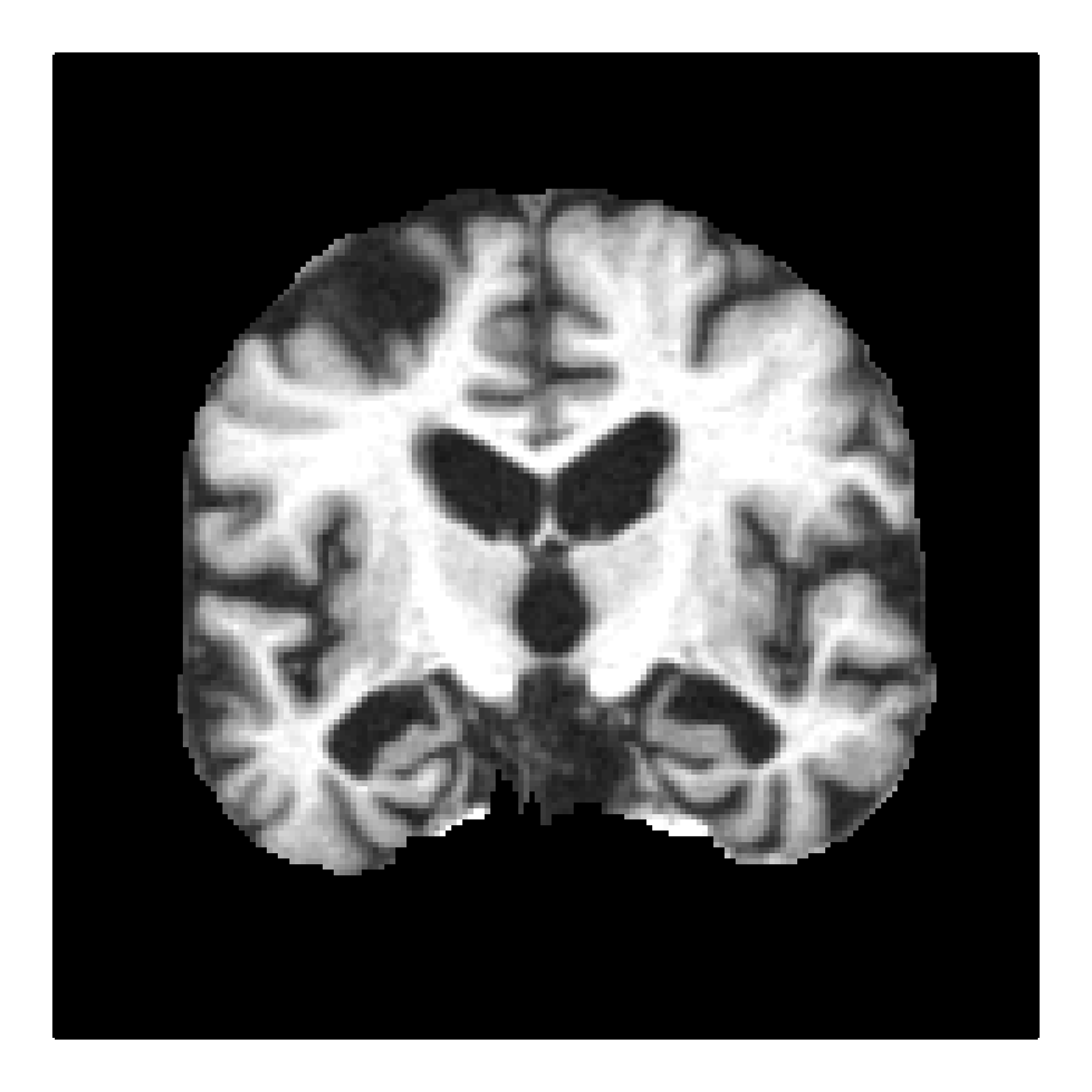}\includegraphics[width=4cm]{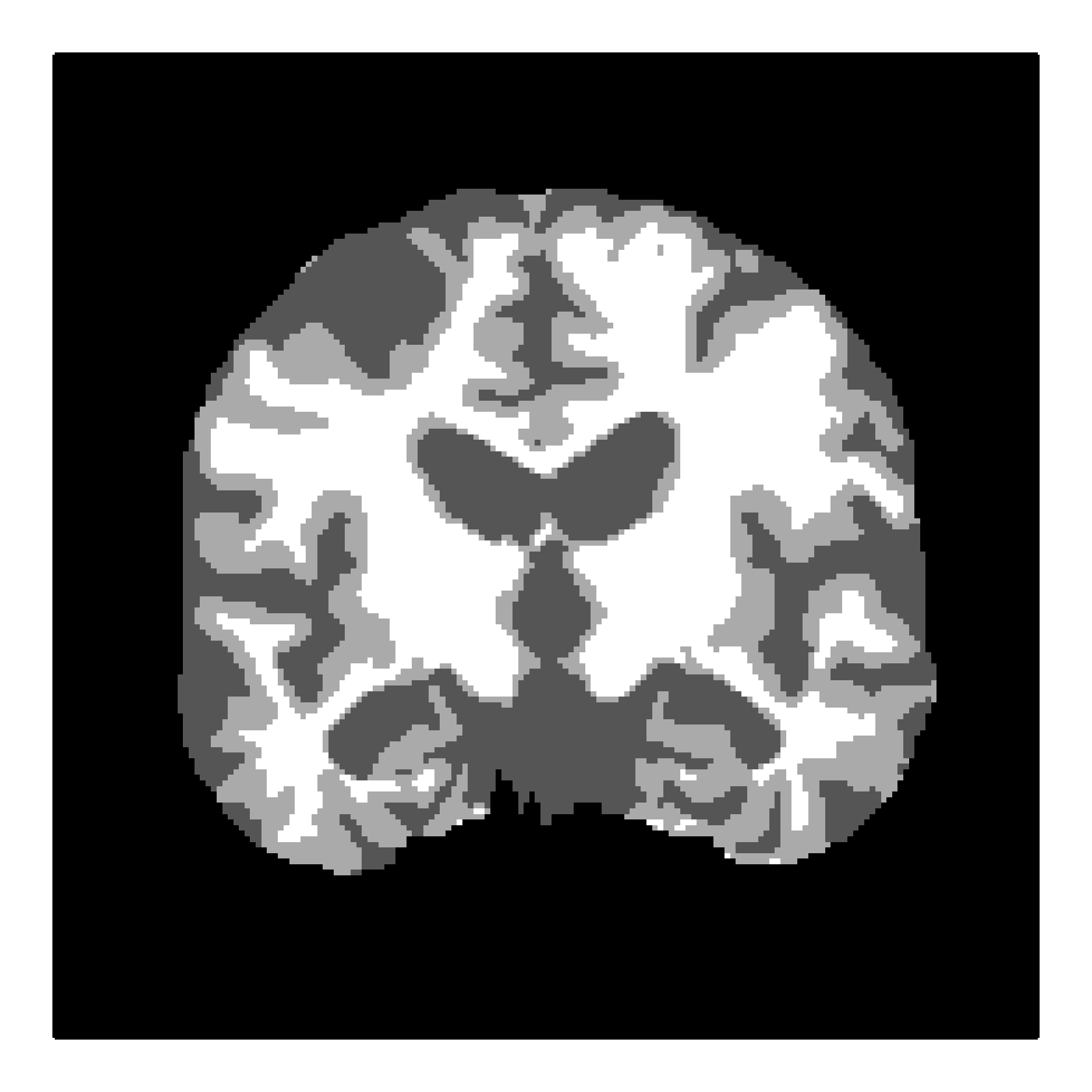}\includegraphics[width=4cm]{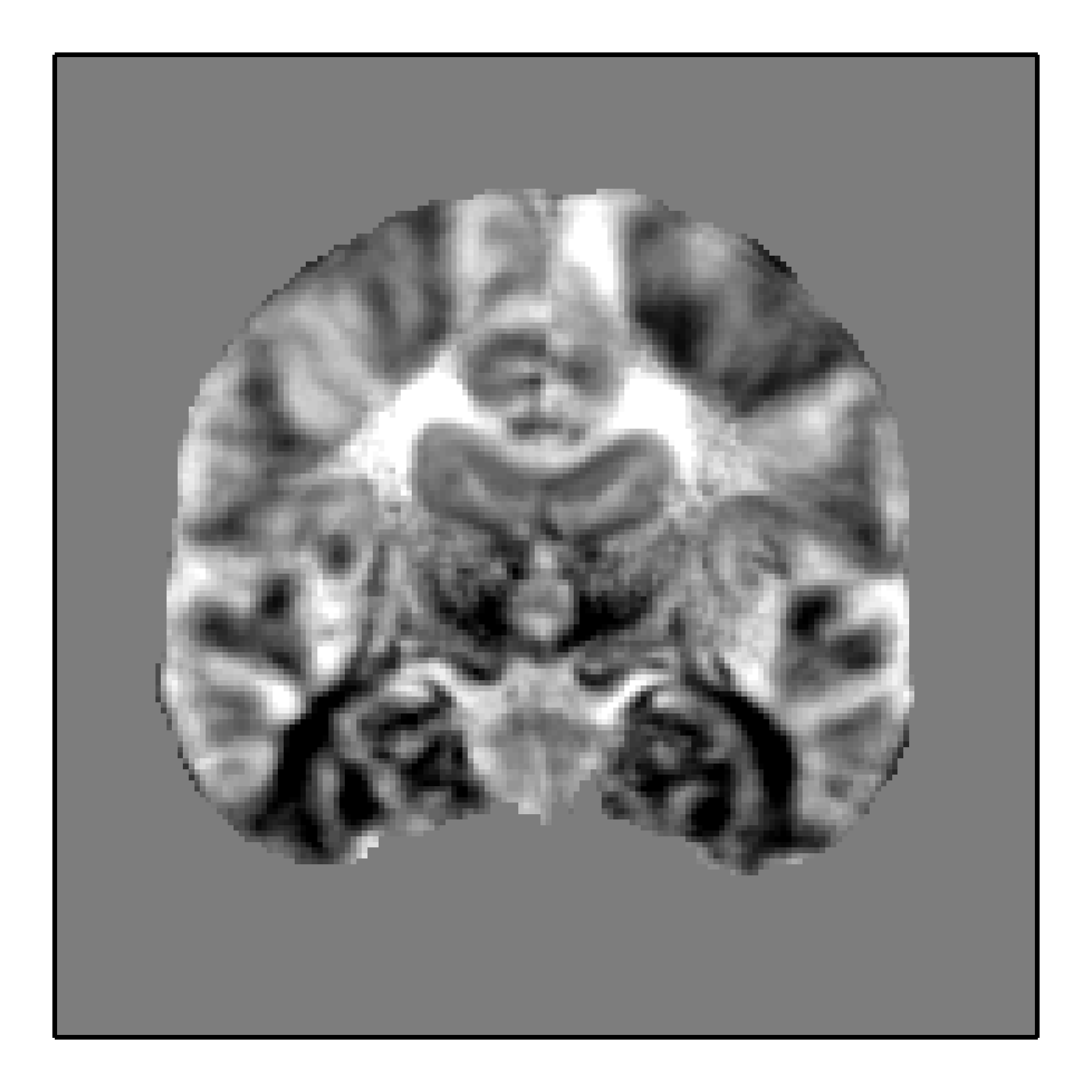}}\medskip{}
\subfloat[Axial raw image, segmented image, and eigenbrain \#7]{\includegraphics[width=4cm]{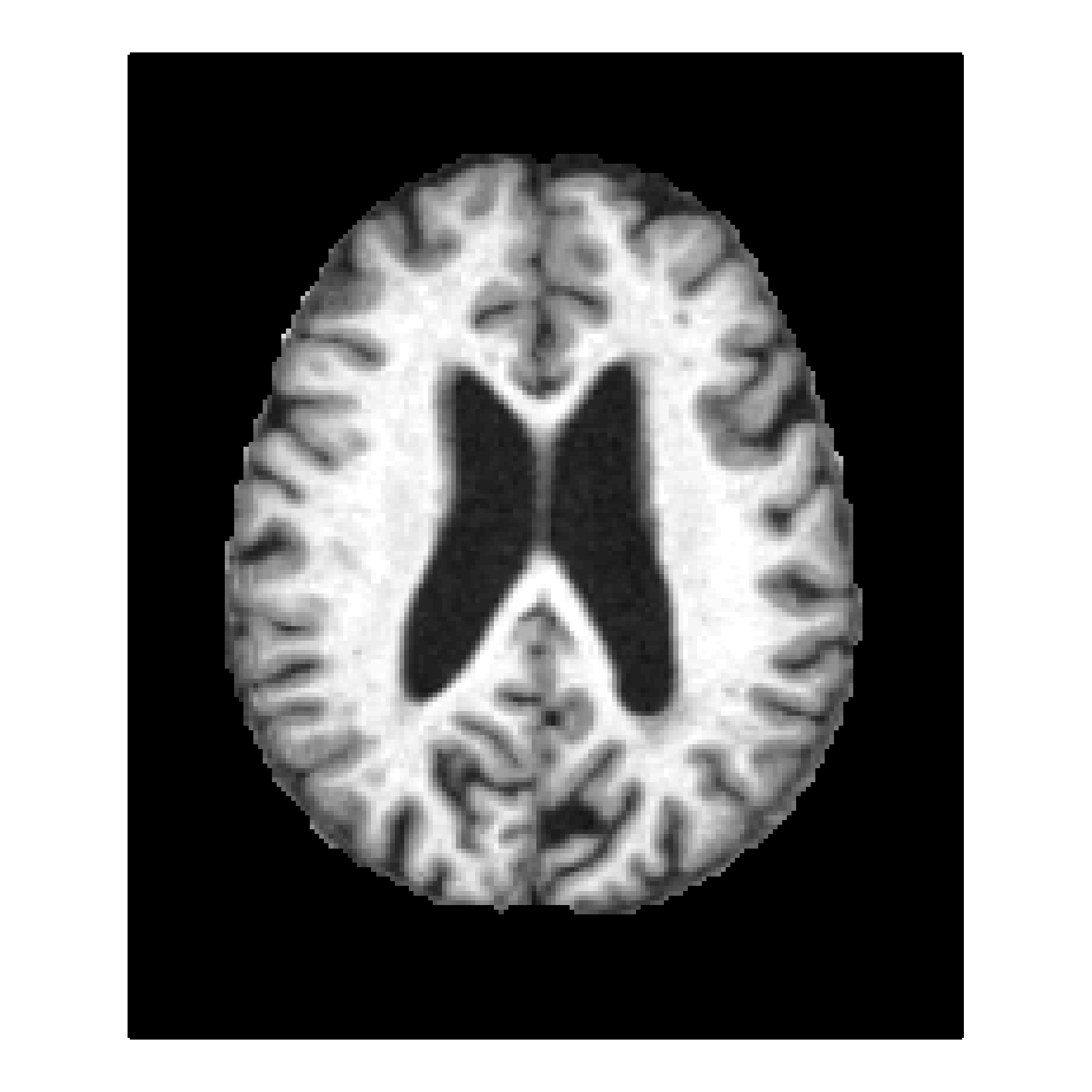}\includegraphics[width=4cm]{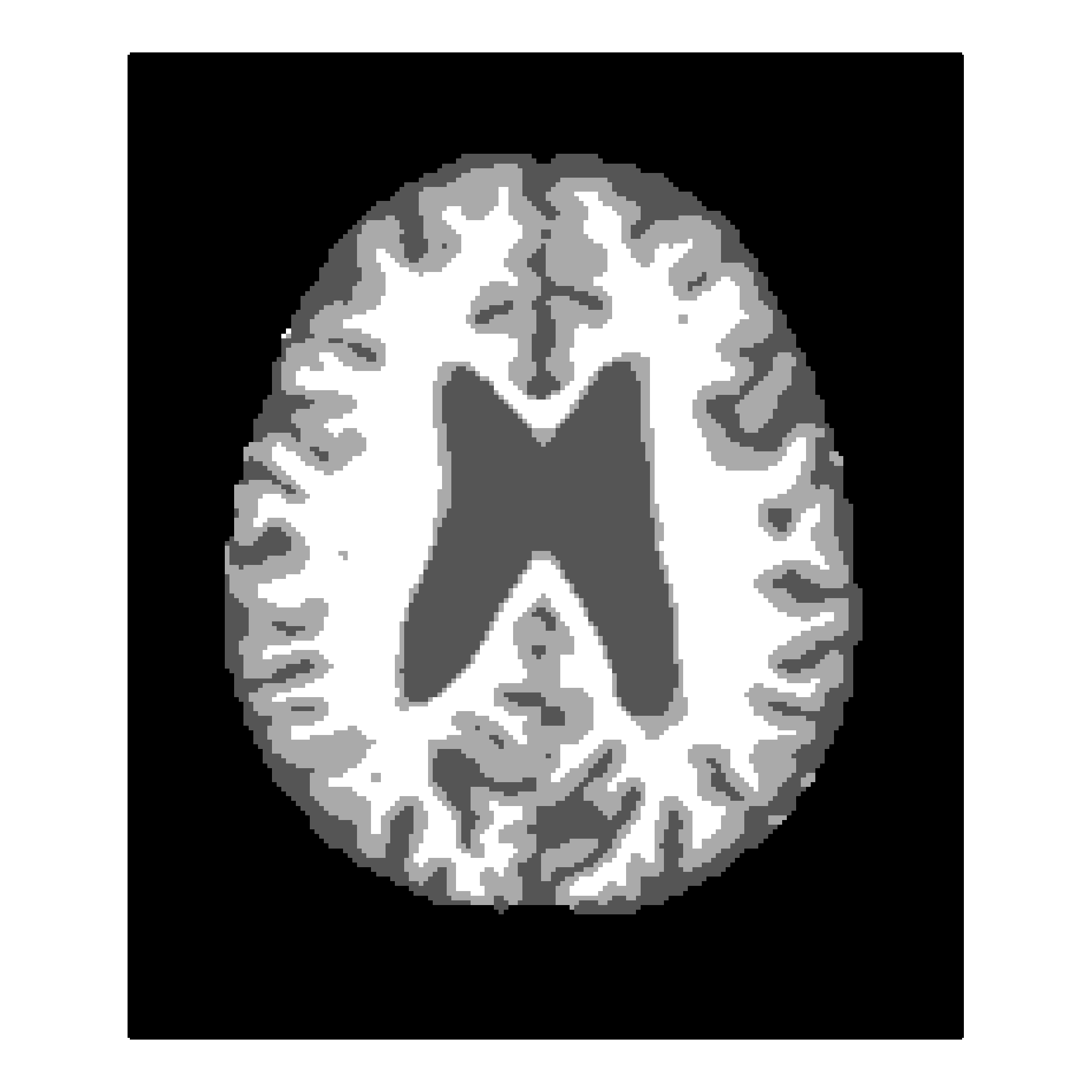}\includegraphics[width=4cm]{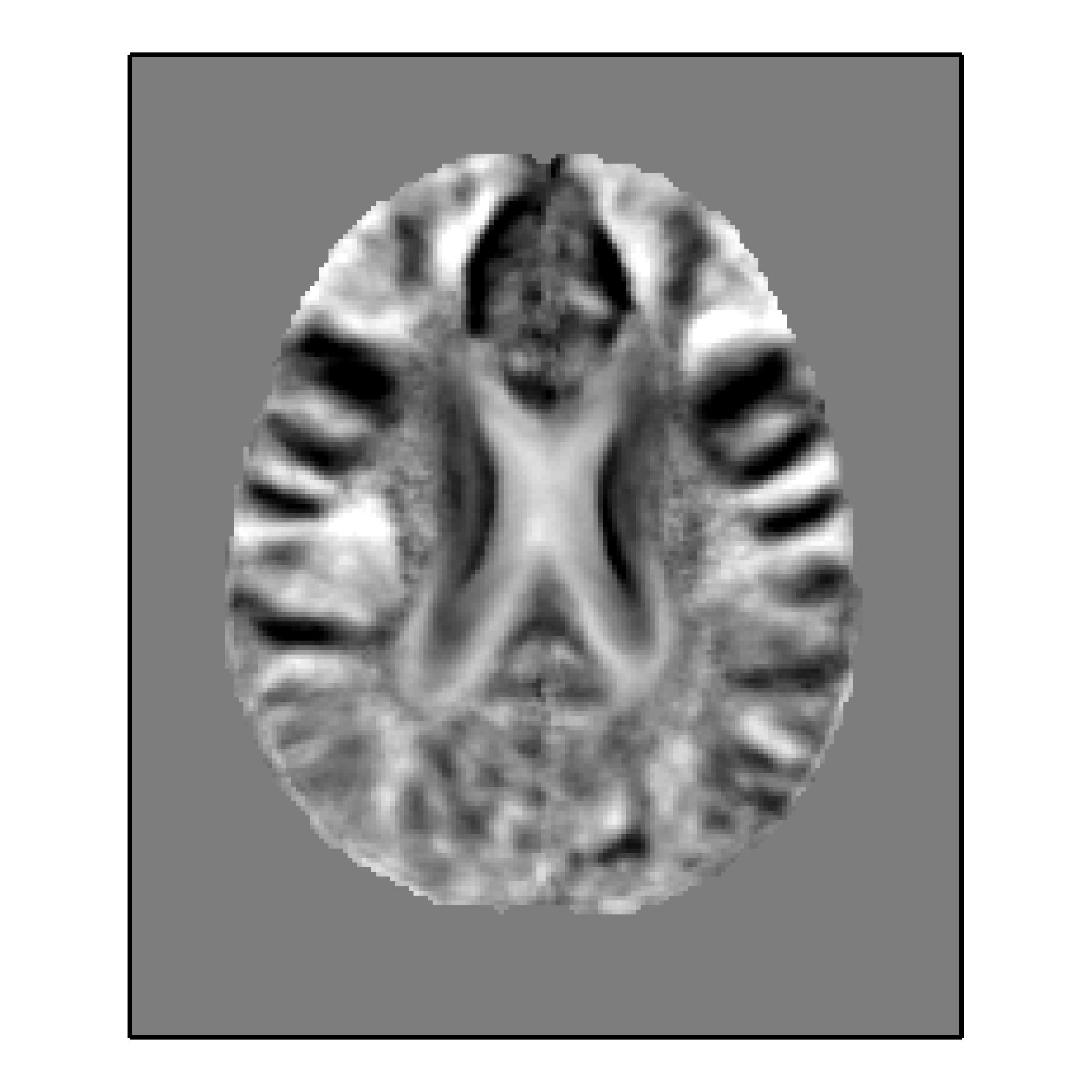}}\caption{\label{fig:brains}}
\end{figure}

We also use principal component analysis (PCA to incorporate more
of the voxel intensity value while keeping the dimensionality to a
minimum. We adopt a strategy well-known for person identification
from facial images, where principal components are computed from a
set of headshots; the resulting principal components are often referred
to as 'eigenfaces' \citep{Turk1991}. Similarly, we compute 'eigenbrains'
for the entire dataset, and consider the component coefficients resulting
from OLS projection of each subject onto the eigenbrain basis set.
We compute eigenbrains for two separate slices, one axial slice and
one coronal slice (figure \ref{fig:brains}), selecting slices on
which gray matter and white matter volume have been shown to correlate
with impaired cognitive ability \citep{Stout1996}. We select PCA
components by parametrically testing the classifers; the combination
of coronal and axial components (\#4 and \#7, respectively) that maximize
test accuracy are shown in figure \ref{fig:brains}.

Using these methods, our feature set includes $11$ values for each
subject, listed in table \ref{tab:List-of-features}. All features
are mean-subtracted and normalized by the variance of each feature.
Lastly, we classify subjects with an CDR score greater than as $0$
as demented, and so $100$ of the $416$ subjects classified with
dementia. 

\begin{table}
\begin{centering}
\caption{\label{tab:List-of-features}List of features included in feature
set}
\begin{tabular}{c|c|c}
\multicolumn{3}{l}{}\tabularnewline
\hline 
\hline 
No. & Feature & Source\tabularnewline
\hline 
\hline 
1 & Age & OASIS\tabularnewline
\hline 
2 & Gender & OASIS\tabularnewline
\hline 
3 & eTIV & OASIS\tabularnewline
\hline 
4 & nBWV & OASIS\tabularnewline
\hline 
5 & Total white matter & $\sum1_{3}(I_{segmented})$\tabularnewline
\hline 
6 & Total gray matter & $\sum1_{2}(I_{segmented})$\tabularnewline
\hline 
7 & Total CSF & $\sum1_{1}(I_{segmented})$\tabularnewline
\hline 
8 & Up/down axial symmetry & Equation \ref{eq:sym_eq}\tabularnewline
\hline 
9 & Left/right axial symmetry & Equation \ref{eq:sym_eq}\tabularnewline
\hline 
10 & Coronal PCA coefficient  & coefficient for component \#4 (coronal)\tabularnewline
\hline 
11 & Axial PCA coefficient &  coefficient for component \#7 (axial)\tabularnewline
\hline 
\end{tabular}
\par\end{centering}

\end{table}

\section{Results}

We compute training accuracy, test accuracy, recall, precision, and
the Matthews correlation coefficient (MCC) to determine performance
of the classifier. an MCC score of $0$ corresponds to random guessing,
and $1$ to perfect classification. The classifiers are tested using
10-fold cross validation; the classifier is trained on $90\%$ of
the dataset, and then tested on the remaining $10\%$ (roughly $40$
subjects), and this is repeated 10 times until every subject has been
tested. The average value of performance metrics (e.g., precision)
is reported here. 

Using more than a single axial and coronal PCA component resulted
in an increase in training accuracy with no corresponding increase
in test accuracy, suggesting the classifier was over fit; therefore,
only a single PCA component was used for the axial and coronal views. 

Using the dimensionally reduced feature set (table \ref{tab:List-of-features}),
the SVM classifier performed with a training accuracy of $86.4\%$
($90.1\%$) and testing accuracy of $85.0\%$ and ($85.7\%$) for
a linear and (gaussian) kernel. The gaussian kernel is a radial basis
function of the form $\exp(-\gamma|u-v|^{2})$, where $\gamma$ was
$\nicefrac{1}{11}$ (or the inverse of the number of features \citep{Chang2011}).
The test precision and recall for each classifier was $68.7\%$ ($68.5\%$)
and $68.0\%$ ($74.0\%$), respectively. The Matthews correlation
coefficient \citep{Baldi2000} for the linear and (gaussian) kernels
was $0.594$ ($0.616$).

In order to ensure the classifiers were not keying off the age feature
alone, classifiers were also tested with a feature set that did not
include age, and test accuracy decreased by $\approx2$\%, indicating
that the classifiers were indeed not solely keying off the age feature.
Additionally, classifiers were run without the PCA components in the
feature set, and recall decreased by $\approx5\%$ ($10\%$) for each
classifier, whereas the precision increased by $\approx2\%$ for both
classifiers; however, the MCC for each classifier excluding PCA components
from the feature set reduced by $.03$ ($.07$), indicating that including
PCA coefficients did improve the performance of the classifiers. 

Previous work \citep{Kloppel2008} achieved testing accuracy of nearly
$90\%$ by using voxel intensity values. Our classifiers do not perform
as well, but there is no evidence of severe over fitting, which may
make a dimensionally reduced feature attractive for classifying truly
out-of-sample subjects.

\section{Conclusions}

We used an SVM with linear and gaussian kernels to classify for a
non-zero CDR score in patients using a dimensionally reduced feature
set. Dimensional reduction was accomplished through image segmentation
and principal component analysis (decomposition into 'eigenbrains').
We achieved MCC values for our classifiers of approximately $0.6$,
and precision and recall of at least $68\%$, with the gaussian kernel
marginally outperforming the linear kernel indicated by the MCC score.
Although performance of the classifier may not equal that of a non-reduced
feature set, we have shown that a dimensionally reduced feature set
does improve the computational tractability, and also does not over
fit the training data, improving the generality of the classification
technique.

\section*{Acknowledgements}

We acknowledge the support for the OASIS project by grants P50 AG05681,
P01 AG03991, R01 AG021910, P50 MH071616, U24 RR021382, R01 MH56584.
Additionally, we would like to acknowledge Professor Andrew Ng for
his lessons and insights into machine learning, and Professor Greg
Zaharchuk for his guidance. 

\bibliographystyle{unsrtnat}
\bibliography{library}

\begin{thebibliography}{12}
\providecommand{\natexlab}[1]{#1}
\providecommand{\url}[1]{\texttt{#1}}
\expandafter\ifx\csname urlstyle\endcsname\relax
  \providecommand{\doi}[1]{doi: #1}\else
  \providecommand{\doi}{doi: \begingroup \urlstyle{rm}\Url}\fi

\bibitem[Marcus et~al.(2007)Marcus, Wang, and Parker]{Marcus2007}
DS~Marcus, TH~Wang, and J~Parker.
\newblock {Open Access Series of Imaging Studies (OASIS): cross-sectional MRI
  data in young, middle aged, nondemented, and demented older adults}.
\newblock \emph{Journal of Cognitive \ldots}, 2007.
\newblock URL
  \url{http://www.mitpressjournals.org/doi/abs/10.1162/jocn.2007.19.9.1498}.

\bibitem[Kl\"{o}ppel et~al.(2008)Kl\"{o}ppel, Stonnington, Barnes, Chen, Chu,
  Good, Mader, Mitchell, Patel, Roberts, Fox, Jack, Ashburner, and
  Frackowiak]{Kloppel2008}
Stefan Kl\"{o}ppel, Cynthia~M Stonnington, Josephine Barnes, Frederick Chen,
  Carlton Chu, Catriona~D Good, Irina Mader, L~Anne Mitchell, Ameet~C Patel,
  Catherine~C Roberts, Nick~C Fox, Clifford~R Jack, John Ashburner, and Richard
  S~J Frackowiak.
\newblock {Accuracy of dementia diagnosis: a direct comparison between
  radiologists and a computerized method.}
\newblock \emph{Brain : a journal of neurology}, 131\penalty0 (Pt 11):\penalty0
  2969--74, November 2008.
\newblock ISSN 1460-2156.
\newblock \doi{10.1093/brain/awn239}.
\newblock URL
  \url{http://www.pubmedcentral.nih.gov/articlerender.fcgi?artid=2577804\&tool=pmcentrez\&rendertype=abstract}.

\bibitem[Mwangi et~al.(2012)Mwangi, Ebmeier, Matthews, and Steele]{Mwangi2012}
Benson Mwangi, Klaus~P Ebmeier, Keith Matthews, and J~Douglas Steele.
\newblock {Multi-centre diagnostic classification of individual structural
  neuroimaging scans from patients with major depressive disorder.}
\newblock \emph{Brain : a journal of neurology}, 135\penalty0 (Pt 5):\penalty0
  1508--21, May 2012.
\newblock ISSN 1460-2156.
\newblock \doi{10.1093/brain/aws084}.
\newblock URL \url{http://www.ncbi.nlm.nih.gov/pubmed/22544901}.

\bibitem[Chang and Lin(2011)]{Chang2011}
Chih-Chung Chang and Chih-Jen Lin.
\newblock {LIBSVM: A Library for support vector machines}.
\newblock \emph{ACM Transactions on Intelligent Systems and Technology},
  2\penalty0 (3):\penalty0 1--27, 2011.
\newblock URL \url{http://www.csie.ntu.edu.tw/~cjlin/libsvm}.

\bibitem[Rubin et~al.(1998)Rubin, Storandt, and Miller]{Rubin1998}
EH~Rubin, M~Storandt, and JP~Miller.
\newblock {A prospective study of cognitive function and onset of dementia in
  cognitively healthy elders}.
\newblock \emph{Archives of \ldots}, 1998.
\newblock URL \url{http://archneur.ama-assn.org/cgi/reprint/55/3/395.pdf}.

\bibitem[Morris(1993)]{Morris1993}
JC~Morris.
\newblock {The Clinical Dementia Rating (CDR): current version and scoring
  rules.}
\newblock \emph{Neurology; Neurology}, 1993.
\newblock URL \url{http://psycnet.apa.org/psycinfo/1994-19989-001}.

\bibitem[Buckner(2004)]{Buckner2004}
RL~Buckner.
\newblock {A unified approach for morphometric and functional data analysis in
  young, old, and demented adults using automated atlas-based head size
  normalization:}.
\newblock \emph{Neuroimage}, 2004.
\newblock URL \url{http://www.miba.auc.dk/~06gr1088d/artikler/Pdf/A unified
  approach for morphometric and functional data analysis in..
  Normalization.pdf}.

\bibitem[Fotenos et~al.(2005)Fotenos, Snyder, and Girton]{Fotenos2005}
AF~Fotenos, AZ~Snyder, and LE~Girton.
\newblock {Normative estimates of cross-sectional and longitudinal brain volume
  decline in aging and AD}.
\newblock \emph{Neurology}, 2005.
\newblock URL \url{http://www.neurology.org/content/64/6/1032.short}.

\bibitem[Zhang et~al.(2001)Zhang, Brady, and Smith]{Zhang2001}
Y~Zhang, M~Brady, and S~Smith.
\newblock {Segmentation of brain MR images through a hidden Markov random field
  model and the expectation-maximization algorithm}.
\newblock \emph{Medical Imaging, IEEE \ldots}, 2001.
\newblock URL
  \url{http://ieeexplore.ieee.org/xpls/abs\_all.jsp?arnumber=906424}.

\bibitem[Turk and Pentland(1991)]{Turk1991}
M~Turk and A~Pentland.
\newblock {Eigenfaces for recognition}.
\newblock \emph{Journal of cognitive neuroscience}, 1991.
\newblock URL
  \url{http://www.mitpressjournals.org/doi/abs/10.1162/jocn.1991.3.1.71}.

\bibitem[Stout and Jernigan(1996)]{Stout1996}
JC~Stout and TL~Jernigan.
\newblock {Association of dementia severity with cortical gray matter and
  abnormal white matter volumes in dementia of the Alzheimer type}.
\newblock \emph{Archives of Neurology}, 53\penalty0 (8):\penalty0 742--749,
  1996.
\newblock \doi{10.1001/archneur.1996.00550080056013.}
\newblock URL \url{http://archneur.ama-assn.org/cgi/reprint/53/8/742.pdf}.

\bibitem[Baldi et~al.(2000)Baldi, Brunak, Chauvin, Andersen, and
  Nielsen]{Baldi2000}
Pierre Baldi, S\o~ren Brunak, Yves Chauvin, Claus A~F Andersen, and Henrik
  Nielsen.
\newblock {Assessing the accuracy of prediction algorithms for classification:
  an overview}.
\newblock \emph{Bioinformatics}, 16\penalty0 (5):\penalty0 412--424, 2000.

\end{thebibliography}

\end{document}